\pdfoutput=1

\documentclass[11pt]{article}

\usepackage{emnlp2022}

\usepackage{times}
\usepackage{latexsym}
\usepackage{amsmath}
\usepackage{graphicx}
\usepackage{subcaption}
\usepackage{booktabs}

\usepackage[T1]{fontenc}

\usepackage[utf8]{inputenc}

\usepackage{microtype}
\usepackage{gb4e}
\noautomath

\title{Syntactic Surprisal From Neural Models Predicts, But Underestimates, Human Processing Difficulty From Syntactic Ambiguities}

\author{Suhas Arehalli \\
  Johns Hopkins University \\
  \texttt{suhas@jhu.edu} \\\And
  Brian Dillon \\
  University of Massachusetts, Amherst \\
  \texttt{brian@linguist.umass.edu} \\\AND
  Tal Linzen \\
  New York University \\
  \texttt{linzen@nyu.edu} \\}

\begin{document}
\maketitle
\begin{abstract}

\end{abstract}
Humans exhibit garden path effects: When reading sentences that are temporarily structurally ambiguous, they slow down when the structure is disambiguated in favor of the less preferred alternative. Surprisal theory \cite{Hale2001, Levy2008}, a prominent explanation of this finding, proposes that these slowdowns are due to the unpredictability of each of the words that occur in these sentences. Challenging this hypothesis, \citet{vanSchijndel} find that estimates of the cost of word predictability derived from language models severely underestimate the magnitude of human garden path effects. In this work, we consider whether this underestimation is due to the fact that humans weight syntactic factors in their predictions more highly than language models do. We propose a method for estimating syntactic predictability from a language model, allowing us to weigh the cost of lexical and syntactic predictability independently. We find that treating syntactic predictability independently from lexical predictability indeed results in larger estimates of garden path. At the same time, even when syntactic predictability is independently weighted, surprisal still greatly underestimate the magnitude of human garden path effects. Our results support the hypothesis that predictability is not the only factor responsible for the processing cost associated with garden path sentences.

\section{Introduction}
Readers exhibit \textit{garden path effects}: When reading a temporarily syntactically ambiguous sentence, they tend to slow down when the sentence is disambiguated in favor of the less preferred parse. For example, a participant who reads the sentence fragment
\begin{exe}
    \ex The suspect sent the file \dots \begin{xlist}
        \ex \dots to the lawyer. \label{ex:MV}
        \ex \dots deserved further investigation. \label{ex:RR}
    \end{xlist}
\end{exe}
can construct a partial parse in at least two distinct ways: In one reading, the verb \textit{sent} acts as the main verb of the sentence, and the continuation of the sentence as an additional argument to \textit{sent} (as in \ref{ex:MV}). In another, less likely, reading, \textit{sent the file} acts as a modifier in a complex subject, which then requires an additional verb phrase to form a complete sentence (as in~\ref{ex:RR}). Prior work has demonstrated that regions like \textit{deserved further investigation}, which disambiguate these temporarily ambiguous sentences in favor of the modifier parse (\ref{ex:RR}), are read slower than those same words would be in an unambiguous version of sentence, such as the following:
\begin{exe}
    \ex The suspect \textit{who was} sent the file deserved further investigation.\label{ex:unambigmvrr}
\end{exe}
In (\ref{ex:unambigmvrr}), the presence of \textit{who was} signals to the reader that \textit{sent the file} acts as a modifier \citep{Frazier1978}. 

One account of this phenomenon, surprisal theory \citep{Hale2001, Levy2008}, suggests that readers maintain a probabilistic representation of all possible parses of the input as they process the sentence incrementally. Processing difficulty in garden path sentences is the cost associated with updating this representation; this cost is proportional to the negative log probability, or surprisal, of the newly observed material under the reader's model of upcoming words. This theory predicts that the slowdown associated with garden path sentences can be entirely captured by the differences in surprisal between the disambiguating region in ambiguous garden path sentences and that same region in a matched unambiguous sentence.

\Citet{vanSchijndel} tested this hypothesis. They estimated the surprisals associated with garden path sentences using LSTM language models (LMs) trained over large natural language corpora. Based on the core assumption of surprisal theory---that processing difficulty on a word, when all lexical factors are kept constant, stands in a constant proportion to the word's surprisal, regardless of its syntactic context---they estimated a conversion factor between surprisal and reading times from non-garden path sentences. Applying this conversion factor to the critical words in garden path sentences, \citeauthor{vanSchijndel} found that surprisal theory, when paired with the surprisals estimated by their models, severely underestimated the magnitude of the garden path effect for three garden path constructions, consistent with attempts to estimate the magnitude of other syntactically-modulated effects \cite{wilcox-etal-2021-targeted}. Moreover, the predicted reading times did not correctly predict differences across the difference garden path constructions, suggesting that no single conversion factor between surprisal and reading times could predict the magnitude of the garden path effect in all three constructions. 

The underestimation documented by \citeauthor{vanSchijndel} can be interpreted in one of two ways: Either (1) surprisal theory cannot, on its own, account for garden path effects; or (2) predictability estimates derived from LSTM LMs fail to capture some aspect of human prediction that is crucial to explaining the processing of garden path sentences. This work investigates the latter possibility. We ask if the gap between the magnitude of human garden path effects in humans and the magnitude that surprisal theory predicts from LMs is due to a mismatch between how humans and LMs weigh two contributors to word-level surprisal: syntactic and lexical predictability. We hypothesize that the LM next-word prediction objective does not sufficiently emphasize the importance that syntactic structure carries for human readers, who may be more actively concerned with interpreting the sentence. In this scenario, since garden paths are the product of unpredictable syntactic structure---as opposed to an unpredictable lexical item---using a LM predictability estimate for the next word could lead to underestimation of garden path effects.
\begin{figure}[t]
    \centering
    \includegraphics[height=2in]{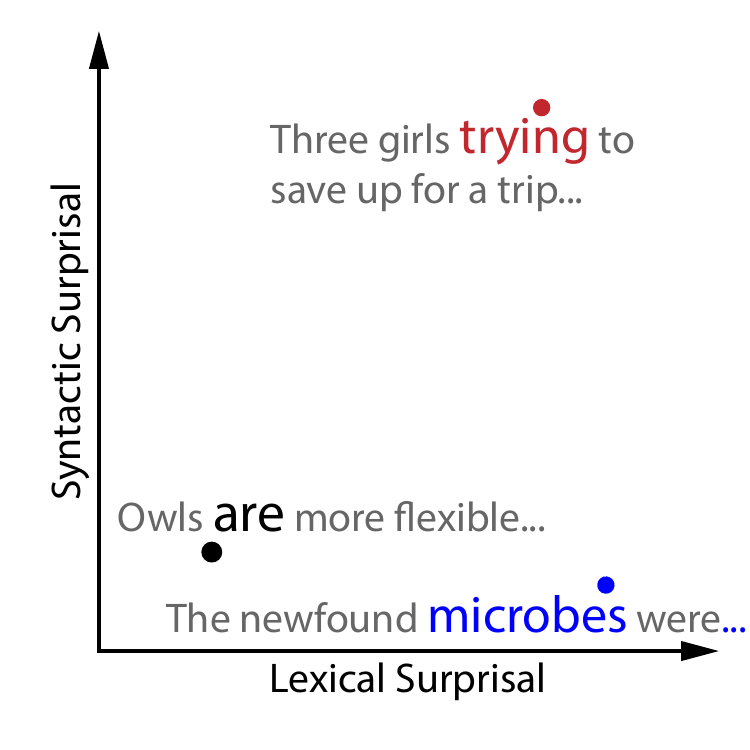}
    \caption{A depiction of the relationship between syntactic and lexical surprisal. Some word tokens, such as \textit{are} in the context of \textit{owls are more flexible}, are highly predictable in all respects. Others are unpredictable due to the syntactic structures they imply (\textit{trying} in \textit{girls trying to save up}), and are expected to be assigned high syntactic and lexical surprisal. Tokens such as \textit{microbes} in the context \textit{the newfound microbes were}, on the other hand, appear in a predictable syntactic environment, but are unpredictable due to their low lexical frequency; such words should be assigned low syntactic surprisal but high lexical surprisal. Since words that appear in unpredictable syntactic environments are themselves unpredictable, we do not expect to find words with high syntactic surprisal but low lexical surprisal.}
    \label{fig:idealized}
\end{figure}

We test the hypothesis that the gap between model and human effects can be bridged by teasing apart the overall predictability of a word from the surprisal associated with the syntactic structure implied by the word (see Figure~\ref{fig:idealized}) and weighting the two factors independently, possibly assigning a higher weight to syntactic surprisal. In this reasoning, we follow prior work on syntactic or unlexicalized surprisal carried out in the context of symbolic parsers, where the probability of a structure and particular lexical item can be explicitly disentangled \cite{Demberg2008, Roark2009}. But while past work has demonstrated that that unlexicalized suprisal from symbolic parsers correlates with measures of human processing difficulty \cite{Demberg2008}, simple recurrent neural networks trained to predict sequences of part-of-speech tags have been shown to track processing difficulty even more strongly \cite{Frank2009}, suggesting that even fairly limited syntactic representations like part-of-speech tags can act as a reasonable proxy of syntactic structure when modeling human behavior.

To compute LSTM-based syntactic surprisal, we train the LM with an auxiliary objective---estimating the likelihood of the next word's supertag under the Combinatory Categorial Grammar (CCG) framework \citep{Steedman1987}---following \citet{enguehard-etal-2017-exploring}. Such supertags can be viewed as enriched part-of-speech tags that encode syntactic information about how a particular word can be combined with its local environment. We then define syntactic surprisal in terms of the likelihood of the next word's CCG supertag, and propose a method of estimating that likelihood using our modified LMs. We validate our formulation of syntactic surprisal by demonstrating that it captures syntactic processing difficulty in garden path sentences, while, crucially, not tracking unpredictability that is due to low frequency lexical items. Following \citet{vanSchijndel}, we then use the syntactic and lexical surprisal values derived from those models to predict reading times for three types of garden path sentences. We find that adding syntactic surprisal as a separate predictor does lead to larger estimates of garden path effects, but those estimates are still an order of magnitude lower than empirical garden path effects. Finally, we discuss the implications of these findings for surprisal theory and single-stage models of syntactic processing.

\section{Computing Syntactic Surprisal}\footnote{Code necessary to reproduce our experiments can be found at \url{https://github.com/SArehalli/SyntacticSurprisal}}
Each incoming word can cause an adjustment in the reader's beliefs about the syntactic structure of the sentence; when a syntactic structure that was assigned a low probability prior to reading the word now has high probability, the word can be said to have high syntactic surprisal. We will operationalize this intuition as the predictability of next word's supertag under the Combinatory Categorial Grammar (CCG) formalism \citep{Steedman1987}: 
\begin{equation}
    \text{surp}_{\text{syn}} = -\log(P(c_{n} \mid w_{1}, ..., w_{n-1})),
\end{equation}
where $c_n$ is the CCG supertag of the $n$-th word. A CCG supertag encodes how a word combines syntactically with adjacent constituents. For example, a token with the tag S\textbackslash NP combines with an NP to its left to form an S constituent, and a token with the tag (S\textbackslash NP)/NP combines with an NP to its right to form an S\textbackslash NP constituent. Since the sequence of supertags associated with all of the words of a sentence often allows only one valid parse, accurately predicting a sentence's supertags has been described as ``almost parsing" \citep{Bangalore1999}; consequently, incremental CCG supertagging can be seen as almost \textit{incremental} parsing.%

We contrast this syntactic surprisal measure with the standard token surprisal measure, which we refer to as \textit{lexical} surprisal:
\begin{equation}
    \text{surp}_{\text{lex}} = -\log(P(w_{n} \mid w_1, ..., w_{n-1})).
\end{equation}
Note that what we call lexical surprisal captures \textit{all} factors that contribute to a token's predictability, including syntactic ones.

In order to compute syntactic and lexical surprisal for a given word, we need models that predict, given a left context, not only the next token, as a standard LM does, but also the next token's supertag. To do this, we train  models with both a language modeling and CCG supertagging objective, and estimate the distribution over the next word's tag by marginalizing over the distribution over the next word that is defined by the LM. Formally, for a sequence of words $w_1, ..., w_n \in W$ with supertags $c_1, ..., c_n \in C$, our model estimates the probability of the next word given all observed words, $p_{w_{n+1}} = P(w_{n+1} \mid w_1, ..., w_n)$, and the probability of the most recent word's supertag given all currently observed words, $p_{c_n \mid w_n} = P(c_n \mid w_1, ..., w_n)$. We then infer the distribution over the next word's supertag as
\begin{align}
    P(c_{n+1} | w_1, ..., w_n)  = \sum_{w^*_{n+1} \in W} p_{c_{n+1} \mid w^*_{n+1}} p_{w^*_{n+1}} \label{eq:pnextword}
\end{align}

If we knew the supertag of the next word $c_{n+1}$, we could simply compute the surprisal of that supertag, $-\log P(c_{n+1} \mid w_1,...,w_n)$. By contrast with lexical surprisal, however---where there is no uncertainty about the identity of $w_{n+1}$ once that word has been read---a word's supertag is often ambiguous during incremental processing. Consider the verb \textit{gathered} in the following sentences, for example:
\begin{exe}
    \ex The squirrels gathered near the tree. \label{ex:sq_intrans}
    \ex The squirrels gathered a few acorns. \label{ex:sq_trans}
\end{exe}
In (\ref{ex:sq_intrans}), \textit{gathered} would eventually be assigned the supertag S\textbackslash NP, indicating that \textit{gathered} is used in its intransitive frame---a number of squirrels assembled together as a group---and takes no direct object. In~(\ref{ex:sq_trans}), on the other hand, the appropriate supertag would be (S\textbackslash NP)/NP, which indicates that in this sentence \textit{gathered} is used in a transitive frame and takes the noun phrase \textit{a few acorns} as a direct object. When processing this sentence incrementally, a reader must maintain this uncertainty over the appropriate supertag for a word past the point at which they have read that word. A measure of syntactic surprisal that aims to model processing difficulty at a particular word should similarly take into account uncertainty over the supertag of a word even after the word itself has been processed. We take this uncertainty into account by using the distribution $p_{c_n \mid w_n}$ that our models define over supertags, and computing syntactic surprisal by marginalizing over this distribution:
\begin{align}
    p_{c_{n+1} \mid w_n} &= P(c_{n+1} \mid w_1, ..., w_n) \\
    surp_{syn} &= -\log \sum_{c_{n+1}^* \in C} p_{c_{n+1}^* \mid w_n} p_{c_{n+1} \mid w_{n+1}}
\end{align}
\subsection{Model Architecture and Training}\
We trained four models, differing only in their random seed, on both a language modeling and CCG supertagging objective. The models consisted of an LSTM shared across the two objectives, which we refer to here as the encoder, and two classifiers, one for language modeling and another for CCG supertagging, which we refer to as the decoders. %

Following \citet{Gulordava2018}, the encoder was a two-layer LSTM with 650 units per layer. Each decoder consists of a single linear layer followed by the softmax operation. For the supertagging objective, we trained using CCGBank \cite{Hockenmaier2007}, a set of CCG annotations for the Wall Street Journal section of the Penn Treebank \cite{Marcus1993}. The corpus we used for language modeling was a concatenation of the Wall Street Journal portion of the Penn Treebank and the 80 million words from the English Wikipedia used by \citet{Gulordava2018}. Language modeling and supertagging losses were weighted equally during training.

Models achieved language modeling perplexities ranging from 74.76 to 75.70 on the test portion of the \citet{Gulordava2018} corpus, while \citet{Gulordava2018}'s best language model achieved a perplexity of 52.0. Models assigned the highest likelihood to the correct CCG supertag in the CCGBank test set between 84.1\% and 84.5\% of the time, compared to bi-LSTM supertaggers which can achieve an accuracy of 94.1\% \cite{vaswani-etal-2016-supertagging}. Note that these supertagging numbers are not directly comparable, as our models use unidirectional LSTMs and thus have no access to a word's right context when supertagging.

\subsection{Experimental data}
We evaluated our model on a subset of the Syntactic Ambiguity Processing (SAP) Benchmark \cite{HuangTBD}, a dataset containing self-paced reading times from 2000 native English speakers who read a variety of syntactically complex constructions as well as comparatively simple filler sentences. The large size of the dataset allows us to get precise estimates of the magnitude of the garden path effect for each of the three types of garden path sentences it contains. We describe each of these constructions in what follows.

\paragraph{Main Verb/Reduced Relative (MVRR)} This ambiguity is illustrated in (\ref{ex:mvrr_amb}):

\begin{exe}
    \ex The suspect sent the file \textbf{deserved} further investigation given the new evidence. \label{ex:mvrr_amb}
    \ex The suspect who was sent the file \textbf{deserved} further investigation given the new evidence. \label{ex:mvrr_un}
\end{exe}

In (\ref{ex:mvrr_amb}), before reading the word \textit{deserved}, the reader can interpret \textit{sent the file} either as a main verb and direct object (where the subject has sent the file) or as a reduced relative clause (where the subject has had the file sent to them). This is disambiguated in favor of the reduced relative clause reading by the next word, \textit{deserved}, which is the true main verb of the complete sentence. We can measure the processing difficulty incurred by this disambiguation by comparing the reading times at \textit{deserved} in (\ref{ex:mvrr_amb}) with the reading times at \textit{deserved} in (\ref{ex:mvrr_un}), where the relative clause \textit{who was sent the file} is unreduced and thus unambiguous.

\paragraph{Noun Phrase/Sentence (NPS)} The NPS ambiguity is illustrated in (\ref{ex:nps_amb}):
\begin{exe}
    \ex The suspect showed the file \textbf{deserved} further investigation during the murder trial. \label{ex:nps_amb}
    \ex The suspect showed that the file \textbf{deserved} further investigation during the murder trial. \label{ex:nps_un}
\end{exe} Before reading \textit{deserved} in (\ref{ex:nps_amb}), \textit{the file} can be interpreted as either a direct object, where the suspect is presenting a file to someone, or as the beginning of a sentential complement, where the suspect is making a point. The word \textit{deserved} disambiguates the sentence in favor of the less frequent sentential complement reading. As before, the matched control sentence (\ref{ex:nps_un}) avoids the ambiguity, here by using the explicit complementizer \textit{that} before \textit{the file}; this control makes it possible to measure the slowdown associated with disambiguation.
\begin{figure*}[t]
    \centering
    \includegraphics[width=\textwidth]{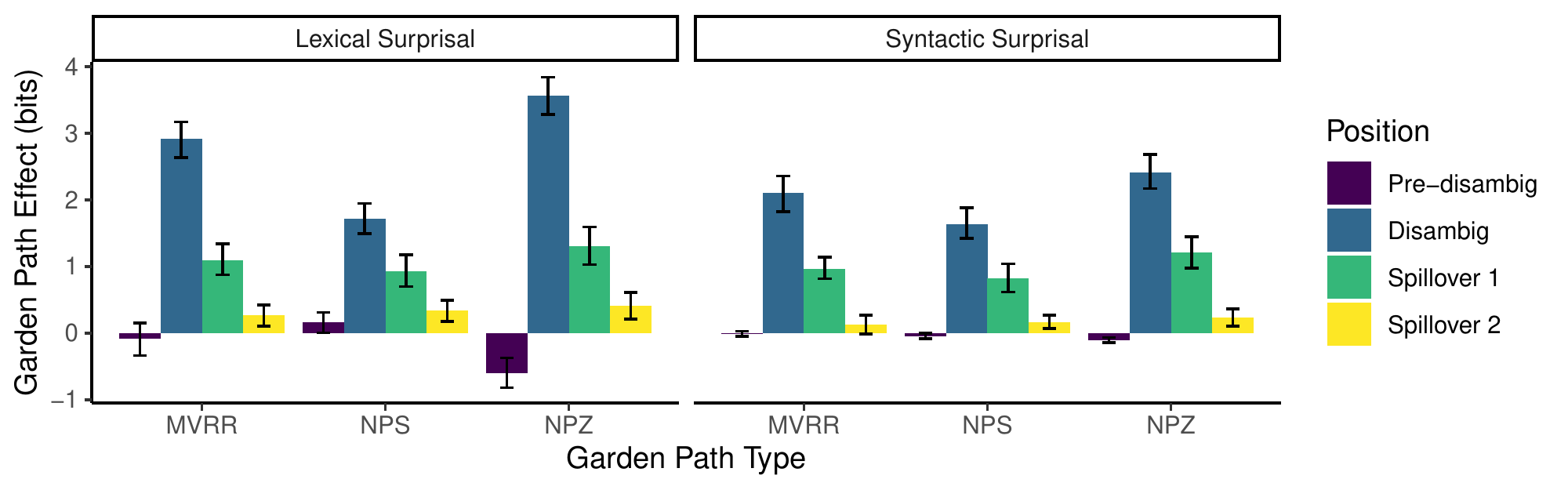}
    \caption{Differences in surprisal estimates between ambiguous and unambiguous garden path sentences at and around the disambiguating verb. Bars indicate 95\% confidence intervals.}
    \label{fig:surp_gps}
\end{figure*}

\paragraph{Noun Phrase/Zero (NPZ):} Finally, in (\ref{ex:npz_amb}), before reading \textit{deserved}, \textit{changed} can be interpreted in two ways: as a transitive verb taking \textit{the file} as a noun phrase direct object (where the file was changed by the suspect); or as an intransitive verb, with \textit{the file} as the subject of a separate clause (where the suspect was changed):

\begin{exe}
    \ex Because the suspect changed the file \textbf{deserved} further investigation during the jury discussions. \label{ex:npz_amb}
    \ex Because the suspect changed, the file \textbf{deserved} further investigation during the jury discussions. \label{ex:npz_un}
\end{exe}

The word \textit{deserved} disambiguates the sentence in favor of the less frequent intransitive reading. Introducing a comma between the clauses in the matched sentence (\ref{ex:npz_un}) removes the ambiguity.

\section{Validating Syntactic Surprisal}

We first validate that our syntactic surprisal measure successfully isolates syntactic predictability from word predictability. To be satisfied that that is the case, we will require two things be true: first, we expect syntactic surprisal to capture processing difficulty that is the result of syntactic unpredictability; and second, we expect that syntactic surprisal is \textbf{not} redundant with lexical predictability. We will evaluate each of these desiderata in turn.

\subsection{Syntactic Surprisal Captures Syntactic Processing Difficulty} 
To verify that syntactic surprisal can capture syntactic unpredictability, we investigate differences in syntactic surprisal between the ambiguous and unambiguous garden path sentences in \citet{HuangTBD}. Since garden path effects are the result of ambiguity about the syntactic structure of a sentence, a difference in surprisal at the point of disambiguation indicates sensitivity to differences in syntactic predictability. 

We found differences in the expected direction for all three types of garden sentences. This was the case both for lexical surprisal---consistent with prior work \citep{Hale2001, vanSchijndel}---and for syntactic surprisal (Figure~\ref{fig:surp_gps}). We did not find differences in the same direction before the point of disambiguation, indicating that the differences we observe after disambiguation are not a consequence differences in surprisal earlier in the sentence that the LM has not fully recovered from.

\begin{figure*}[t]
    \centering
    \begin{subfigure}{0.45\textwidth}
    \includegraphics[width=\textwidth]{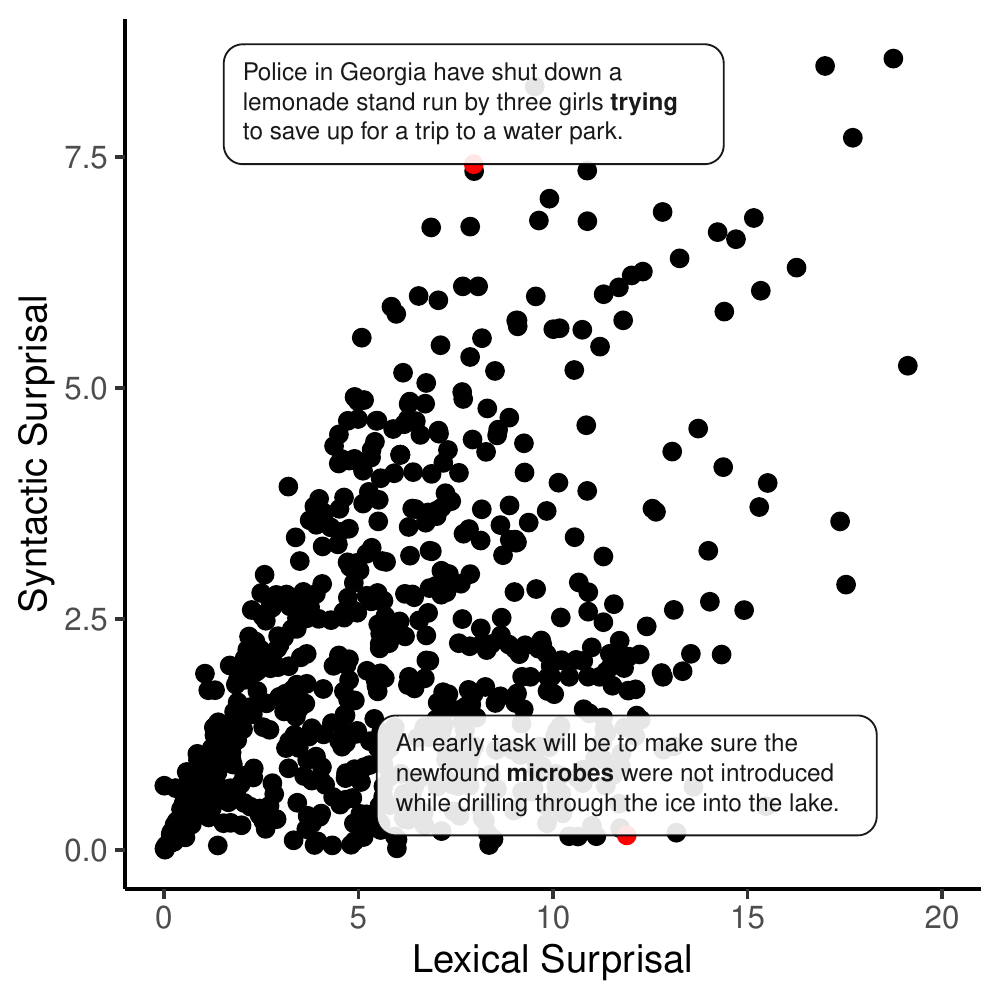}
    \caption{Lexical vs. syntactic surprisal.}
    \label{fig:surp_corr}
    \end{subfigure}\hfill
    \begin{subfigure}{0.45\textwidth}
    \includegraphics[width=\textwidth]{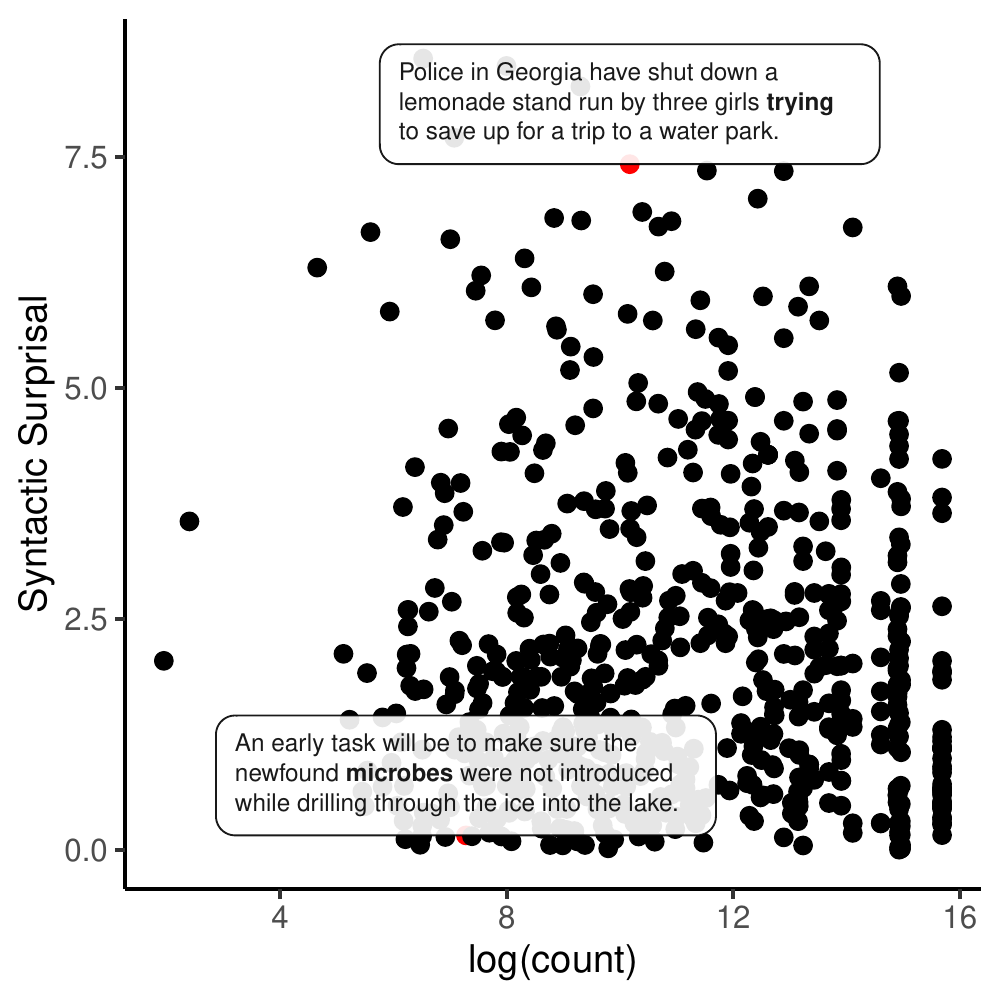}
    \caption{Log unigram frequency vs. syntactic surprisal.}
    \label{fig:surp_freq}
    \end{subfigure}
    \caption{Correlations between syntactic surprisal, lexical surprisal, and unigram frequency for each word in the filler items of \protect\citet{HuangTBD}. Since these results are fairly consistent across model instances, we present results from a single instance. Two words --- one with high syntactic surprisal and high lexical surprisal and one with high lexical surprisal but low syntactic surprisal --- are labeled with their context.}
    \label{fig:corrs}
\end{figure*}

\subsection{Syntactic Surprisal Captures Only Syntactic Predictability}
To verify that syntactic surprisal successfully isolates syntactic factors on predictability, we make two comparisons: first, to lexical surprisal, to verify that syntactic surprisal does not capture all of the variance captured by lexical surprisal; and second, to unigram frequency, to verify that syntactic surprisal is not driven by the frequency of specific lexical items.

\paragraph{Syntactical surprisal does not capture all of the variance captured by lexical surprisal} If syntactic surprisal captures a strict subset of the variance captured by lexical surprisal, we expect to see a subset of words with high lexical surprisal and low syntactic surprisal (in addition, perhaps, to words with highly correlated syntactic and lexical surprisals). This subset should represent words that are unpredictable for reasons that are independent of the syntactic structures they imply. By contrast, words that introduce infrequent syntactic structures should have both high syntactic surprisal and high lexical surprisal, as the unpredictability of the syntactic structure means that a word that implies that structure is necessarily unpredictable. This matches what we see in Figure~\ref{fig:surp_corr}: The relatively frequent verb \textit{trying} introducing a reduced relative clause has high syntactic and lexical surprisal, while infrequent nouns like \textit{microbe} have low syntactic surprisal but high lexical surprisal. 

\paragraph{High syntactic surprisal does not reflect low unigram frequency} In Figure~\ref{fig:surp_freq}, we plot the syntactic surprisals of words from the filler items with their log-frequency in the Corpus of Contemporary American English (COCA; \citealt{Davies2008}). We find no significant correlation between the two ($r=-0.021$, $t=-1.10$, $p = 0.27$), indicating that lexical frequency does not drive syntactic surprisal effects.

These three results --- that syntactic surprisal captures garden path effects, that we find a subset of words with low syntactic surprisal and high lexical surprisal, and that we find no evidence of low lexical frequency driving syntactic surprisal --- suggest that syntactic surprisal captures only the syntactic contributions to a word's unpredictability. We will now use syntactic surprisal in concert with lexical surprisal to directly predict the magnitude of garden path effects. 

\section{Evaluating Against Human Reading Times}
Recall that surprisal theory assumes a linear relationship between surprisal and measures of processing difficulty such as reading times. We follow \citet{vanSchijndel} and estimate a mapping between our surprisal measures and reading times by fitting linear mixed-effects models to the filler (i.e., non-garden path) materials from \citet{HuangTBD}.  In order to compare syntactic and lexical surprisal, we fit four conversion models: one with syntactic surprisal as a predictor, one with lexical surprisal, one with both types of surprisal, and one that does not include either surprisal measure. All four models included baseline predictors other than surprisal --- unigram frequency, word position, and word length --- which on their own are not expected to capture garden path effects. To account for spillover effects, where processing difficulty from a word spills over to affect reading times at future words, we included all of the aforementioned factors (except word position) not only for the current word but also for the two prior words (a simplification of the technique of \citealt{vanSchijndel}). This process is repeated with each of the four sets of surprisals extracted from our four language model/supertagger instances. Further details about the surprisal-to-RT conversion process are presented in Appendix~\ref{app:conversion}.
After all four of our models have been fit to the filler items, we use the estimated coefficients to predict reading times for the each of the critical items. 

\section{Results}
\begin{figure*}[t]
    \centering
    \begin{subfigure}{\textwidth}
    \includegraphics[height=2.2in]{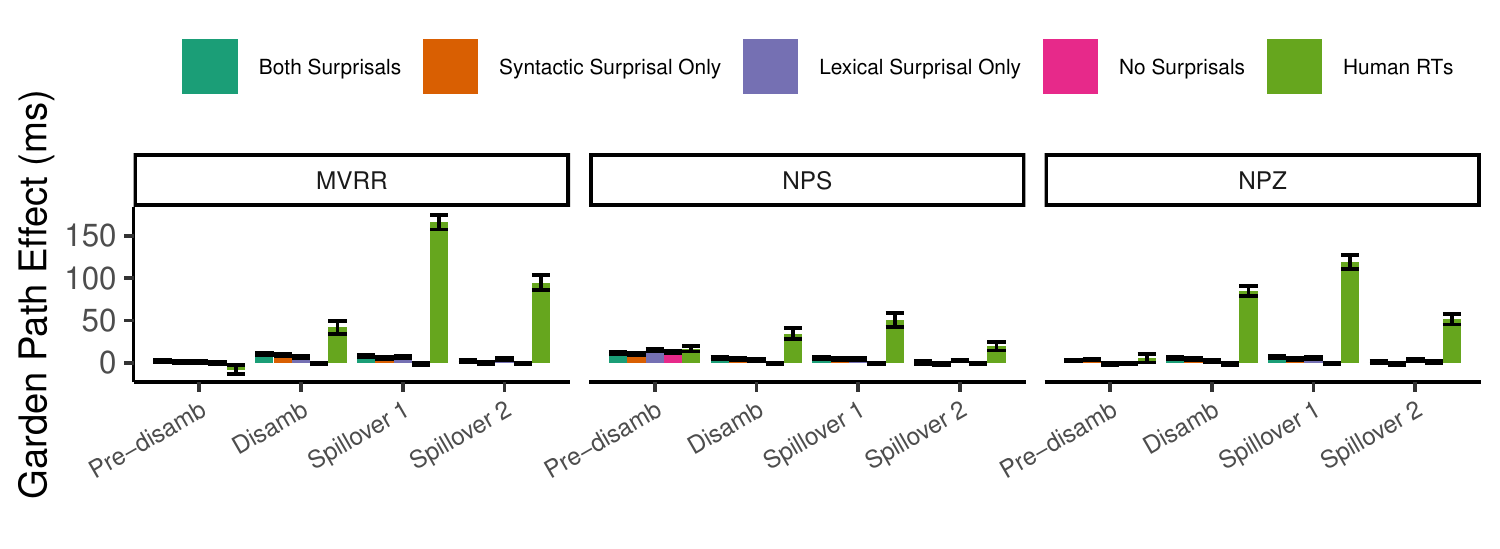}
    \caption{\label{fig:results}Model predictions and human results.}
    \end{subfigure}
    \begin{subfigure}{\textwidth}
        \includegraphics[height=3in]{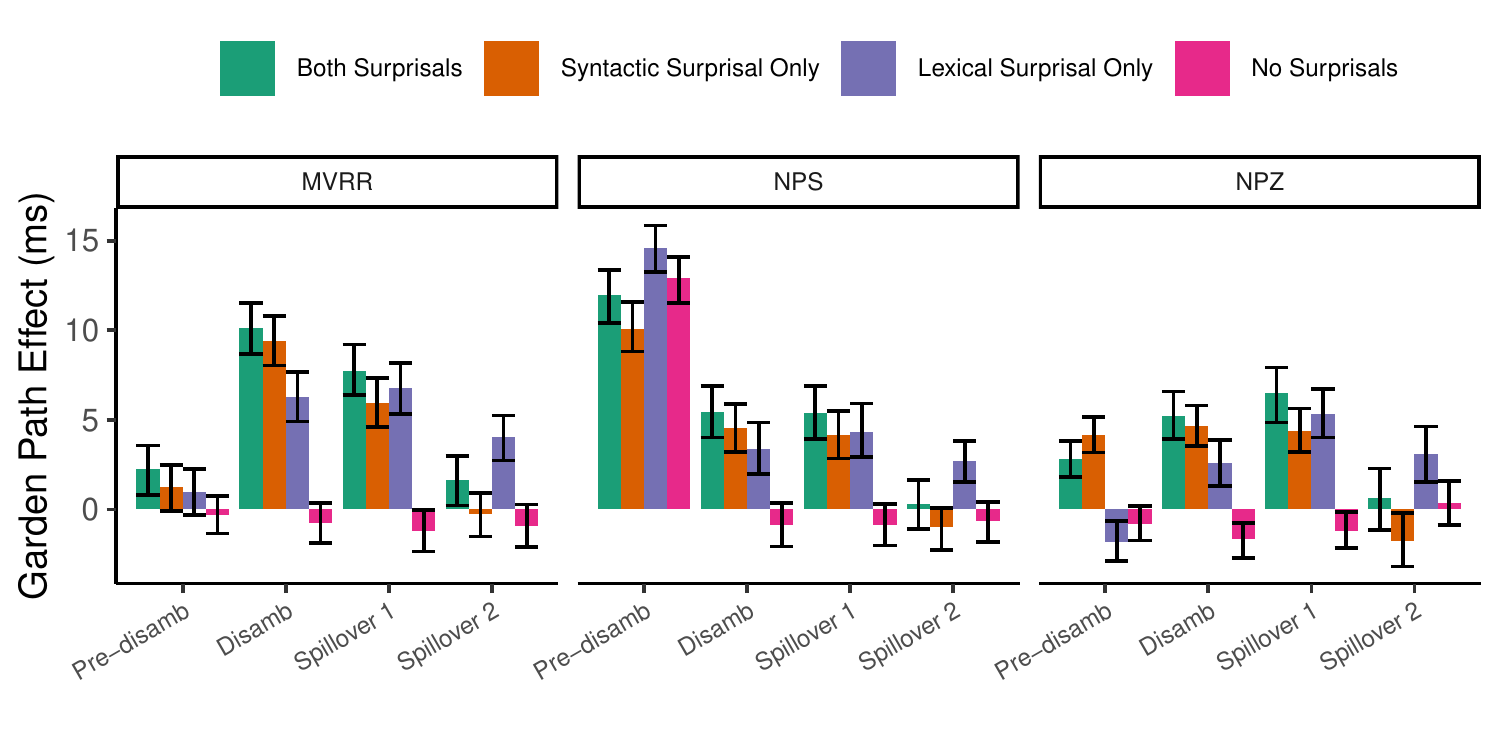}
            \caption{\label{fig:results_model}Model predictions only.}
    \end{subfigure}

    \caption{Empirical and model-predicted readings times for the three garden path constructions. Bars indicate the difference between the mean readings times for the ambiguous and unambiguous sentences across participants for each condition. Error bars indicate bootstrapped 95\% confidence intervals.}
    
\end{figure*}

Predicted RT differences from our conversion models, as well as the RT differences observed in humans, are presented in Figure~\ref{fig:results}. Regardless of the predictors used in the mixed-effects model---lexical surprisal, syntactic surprisal, neither, or both---predicted reading time differences greatly underestimate the reading time differences observed in humans. This is unlikely to be an issue with our surprisal-to-reading-times conversion method more broadly, as at the pre-disambiguation word, RTs and predicted RTs match much more closely than in post-disambiguation regions (particularly in capturing effects in the pre-disambiguation region in NPS sentences), indicating that the difference in magnitudes is due specifically to an underestimation of the garden path effect.

\begin{table*}[t]
    \centering
    \resizebox{\textwidth}{!}{
    \begin{tabular}{lccc}
        \toprule
         Disambig & MVRR & NPS & NPZ \\
         \midrule
         Both vs. Syntactic Only &$\beta = 0.83$, $p < 0.001$ &  $\beta = 0.89$, $p < 0.001$ & $\beta = 0.72$, $p < 0.001$ \\
         Syntactic Only vs. Lexical Only & $\beta = 2.91$, $p < 0.001$ & $\beta = 1.55$, $p < 0.001$ & $\beta = 1.53$, $p < 0.001$\\
         Syntactic Only vs. Neither & $\beta = 10.10$, $p < 0.001$& $\beta = 6.03$, $p < 0.001$ & $\beta = 5.77$, $p < 0.001$\\
         \bottomrule
         Spillover 1 & MVRR & NPS & NPZ \\
         \midrule
         Both vs. Syntactic Only & $\beta = 1.80$, $p < 0.001$ &  $\beta = 1.12$, $p < 0.001$ & $\beta = 2.15$, $p < 0.001$ \\
         Syntactic Only vs. Lexical Only & $\beta = -0.71$, $p < 0.001$ & $\beta = 0.056$, $p = 0.98$ & $\beta = -1.31$, $p < 0.001$\\
         Syntactic Only vs. Neither & $\beta = 7.45$, $p < 0.001$ & $\beta = 5.26$, $p < 0.001$ & $\beta = 5.09$, $p < 0.001$\\
         \bottomrule
         Spillover 2 & MVRR & NPS & NPZ \\
         \midrule
         Both vs. Syntactic Only & $\beta = 1.98$, $p < 0.001$ &  $\beta = 1.29$, $p < 0.001$ & $\beta = 2.29$, $p < 0.001$ \\
         Syntactic Only vs. Lexical Only & $\beta = -4.58$, $p < 0.001$ & $\beta = -3.68$, $p < 0.001$ & $\beta = -4.83$, $p < 0.001$\\
         Syntactic Only vs. Neither & $\beta = -0.51$, $p < 0.01$ & $\beta = -0.062$, $p = 0.96$ & $\beta = -2.62$, $p < 0.001$\\
         \bottomrule
    \end{tabular}
    }
    \caption{Results of a Linear Mixed Effects analysis over our model-predicted reading times for our effect of interest: the interaction between ambiguity and the conversion model. A significant result with a positive coefficient indicates that the conversion model on the left side of the contrast label predicted a significantly larger garden path effect than the model on the right. See Appendix~\ref{app:LMEMpred} for further details.}
    \label{tab:predicted_results}
\end{table*}
While the inclusion of syntactic surprisal does not close the gap between model predictions and the empirical reading times, it does typically lead to a larger predicted garden path effect. To see this difference more clearly, in Figure~\ref{fig:results_model} we exclude the human reading times and zoom in on the garden path effects predicted by the models. 
To determine whether adding syntactic surprisal as a predictor affected the magnitude of the garden path effects we predicted, we fit a Linear Mixed Effects Model over all of our conversion models' predicted reading times for each garden path construction at each word in the critical region. We present the results of this analysis for the effect of interest (the interaction between the conversion model and the garden path effect) in Table~\ref{tab:predicted_results}. We find that (1) models containing both surprisals predicted the largest garden path effects at the disambiguating word and first spillover word, (2) models with only syntactic surprisal predicted greater garden path effects than models with only lexical or no surprisal at the disambiguating word, and (3) models with only lexical surprisal predicted larger garden path effects than models with only syntactic or no surprisal in the spillover regions. Note that while models with only lexical surprisal did predict larger effects than other conversion models at the second spillover word, the fact that this only takes place long after the disambiguating word suggests that this difference is due to differing spillover profiles amongst our surprisal measures. Since this work focuses on the estimation of the magnitude of garden path effects, we leave an investigation of this to future work.
\section{Discussion}
What is the source of the discrepancy between the magnitude of garden path effects in humans and surprisal-based estimates of those magnitudes from neural network language models? In this paper, we have evaluated one possible answer to this question: that word predictability estimates from LMs underweight the importance of syntax to the predictions made by humans. We have proposed a method of estimating syntactic predictability from LSTM LMs augmented with a CCG supertagging auxiliary objective; confirmed that this measure matches our intuitive desiderata from a syntactic surprisal measure; and compared garden path effect magnitude predictions derived from standard, lexical surprisal and syntactic surprisal. Our main finding is that while the syntactic surprisal measure we propose does typically lead to larger predicted garden path effects, model-predicted garden path effects still vastly underestimate the magnitude of garden path effects found in humans.

We defined syntactic surprisal in terms of the predictability of the next word's CCG supertag. This choice is motivated by the relative simplicity of computing this measure---a straightforward auxiliary objective that can be added to any conceivable neural language model---as well as two substantive desiderata: First, we would like the measure to capture processing difficulty due to syntactic unpredictability. Since a word's CCG supertags captures how the word combines with the local syntactic structure, we hypothesize that the surprisal of that supertag---which indicates the extent to which that syntactic combination is unexpected---is a good predictor of syntactic unpredictability. This was borne out in our analysis that showed that syntactic surprisal predicts differences in the correct direction in three garden path constructions. 

Second, since syntactic surprisal is designed to isolate \textit{syntactic} predictability from other forms of predictability, it should \textit{not} be perfectly correlated with lexical factors. The comparisons to lexical surprisal and word frequency showed that this desideratum was met: We were able to identify in our materials words that were lexically surprising but had low syntactic surprisal, and we found a \textit{positive} correlation between frequency and syntactic surprisal --- the opposite of what would be predicted if high syntactic surprisal was driven by low word frequency. %

The increase in model-predicted garden path magnitudes when we use syntactic surprisal, compared to using just standard lexical surprisal, suggests that predictability estimates from LSTM LMs indeed understate the role that syntactic factors play in human prediction. To see why that is, recall that syntactic surprisal captures a subset of the variance that lexical surprisal does. The fact that adding syntactic surprisal produces a better fit to human reading times than lexical surprisal, then, suggests that syntactic factors affect lexical surprisal less than they would need to in order to capture variation in human reading times. One potential explanation for this discrepancy is the difference in the tasks humans and LMs perform: While LMs need only predict words in corpora, humans must to comprehend what they read. While both tasks demand some sensitivity to syntactic structure, the need to interpret sentences may place greater importance on predicting structure, leading to a higher sensitivity to syntactic unpredictability.

While models with syntactic surprisal provided a better fit to the human data than those with just lexical surprisal, there remained a very large discrepancy between model-predicted and human garden path effect sizes. It may be possible to further close this gap within the surprisal framework using different approaches to estimating syntactic predictability; one such approach could rely on Recurrent Neural Network Grammars \cite{Dyer2016}, which derive word-level predictability estimates from explicit syntactic parsing mechanisms. %

Another possibility is that the discrepancy is not due to flaws in our estimates of human predictability: perhaps surprisal, even based on a perfect simulation of human predictions, is simply not the correct account of the magnitude of the garden path effect observed in humans \cite{vanSchijndel}. One family of alternative accounts consists of \textit{two-stage, serial} models of processing \cite{Frazier1978,Fodor1994,Lewis1998,Bader1998,Sturt1999}. In such a model, when readers first read through the ambiguous fragment of the sentence, they commit to a small set of preferred parses. When they reach a disambiguating region where all of the parses they have committed to are no longer consistent with the input, a reader would engage a separate, costly reanalysis process in order to construct a new partial parse consistent with the all of the currently available input. The processing cost associated with this reanalysis procedures incurs a slowdown in reading times that does not occur in an unambiguous sentence where the incorrect initial parse is not available, resulting the garden path effects that we observe. %
Unlike surprisal-based accounts, however, it is often unclear how to derive broad-coverage, quantitative predictions for the size of garden path effects from existing two-stage accounts. As a result, it is difficult to know whether the quantitative mismatches between surprisal-accounts and human reading times that we observed should be taken as evidence for an explicit reanalysis process. This further highlights the need for precise implementations of two-stage serial models that we can quantitatively evaluate against surprisal accounts.

\section*{Acknowledgements}

This work was supported by the National Science Foundation, grant nos. BCS-2020945 and BCS-2020914, by the United States--Israel Binational Science Foundation (award no. 2018284), and in part through the NYU IT High Performance Computing resources, services, and staff expertise. We would also like to thank members of the NYU Computation and Psycholinguistics lab, as well as members of the Society for Human Sentence Processing community, for insightful discussion around this work. 

\bibliography{refs,custom, main}
\bibliographystyle{acl_natbib}
\appendix

\section{Appendix}
\label{sec:appendix}
\subsection{Converting Surprisals to Reading Times}\label{app:conversion}
In order to gauge the impact of syntactic surprisal on the predicted reading time at word $n$, $rt_n$, we fit four mixed effects models over the filler data: one containing only lexical surprisal ($s^{lex}_n$), one containing only syntactic surprisal ($s^{syn}_n$), one containing both, and one containing neither. As reading times are sensitive to other features of the word being read like unigram frequency ($f_n$), position in sentence $p$, and length in characters ($c_n$), we include those variables as additional factors in the regression. In order to account for spillover effects, where processing difficulty from a word often surfaces in the reading times of subsequent words, we include all of the aforementioned factors for the prior two words. We additionally include random intercepts by item and by participant, as well as random slopes by item for all of the surprisal fixed effects. This gives us the following linear mixed effects model formulas:
\begin{align*}
\begin{split}
    rt_n &\sim f_n * c_n + f_{n-1} * c_{n-1} \\
         &+ f_{n-2} * c_{n-2} + p \\
         &+ (1 \mid \text{item}) + (1 \mid \text{participant})
    \end{split} \tag{neither} \\
\end{align*}
\begin{align*}
    \begin{split}
    rt_n &\sim s^{lex}_n + s^{lex}_{n-1} + s^{lex}_{n-2}  \\
         &+ f_n * c_n + f_{n-1} * c_{n-1} \\
         &+ f_{n-2} * c_{n-2} + p \\
         &+ (1 + s^{lex}_n + s^{lex}_{n-1} + s^{lex}_{n-2} \mid \text{item}) \\
         &+ (1 \mid \text{participant})
    \end{split} \tag{lexical} \\
\end{align*}
\begin{align*}
    \begin{split}
    rt_n &\sim s^{syn}_n + s^{syn}_{n-1} + s^{syn}_{n-2} +  \\
         &+ f_n * c_n + f_{n-1} * c_{n-1} \\
         &+ f_{n-2} * c_{n-2} + p \\
         &+ (1 + s^{syn}_n + s^{syn}_{n-1} + s^{syn}_{n-2} \mid \text{item}) \\
         &+ (1 \mid \text{participant}) 
    \end{split} \tag{syntactic} \\
\end{align*}
\begin{align*}
    \begin{split}
    rt_n &\sim s^{lex}_n + s^{lex}_{n-1} + s^{lex}_{n-2} \\
         &+ s^{syn}_n + s^{syn}_{n-1} + s^{syn}_{n-2} \\
         &+ f_n * c_n + f_{n-1} * c_{n-1} \\
         &+ f_{n-2} * c_{n-2} + p  \\
         &+ (1 + s^{lex}_n + s^{lex}_{n-1} + s^{lex}_{n-2} \\
         &+ s^{syn}_n + s^{syn}_{n-1} + s^{syn}_{n-2} \mid \text{item}) \\
         &+ (1 \mid \text{participant})
    \end{split} \tag{both}
\end{align*}
These models were fit using filler data from \citet{HuangTBD}, and the coefficients from each model were used to predict reading times for all of the critical, garden path items from the corresponding surprisals, frequencies, lengths, and positions.
\subsection{Statistical Analysis of Predicted RTs}\label{app:LMEMpred}
To analyze the predicted reading times that come from our four models of surprisal-to-reading time conversion, we fit three separate linear mixed effects models: one over MVRR garden paths, one over NPS garden paths, and one over NPZ garden paths. Each model includes fixed effects of ambiguity and the types of surprisals used in predicting reading times: syntactic surprisal only, lexical surprisal only, both surprisals, or neither. Crucially, we include the interaction between these two factors, representing how our choice of surprisal-to-RT conversion model affects the size of the predicted garden path effect. We additionally include random intercepts by item and by participant. This results in the following mixed effects model formula:
\begin{align*}
    pred\_rt &\sim ambiguity * model \\
              &+ (1 \mid \text{item}) + (1 \mid \text{participant}).
\end{align*}
Since we have four different models converting between surprisals and RTs, we estimate three contrasts for the interaction term: the model with both surprisals vs. the model with only syntactic surprisals, the model with only syntactic surprisals vs. the model with only lexical surprisals, and the model with only lexical surprisals vs. the model with neither surprisal. The estimated magnitude (represented by the $\beta$ coefficient) as well as significance of the difference for each of these contrasts is reported in the main text in Table~\ref{tab:predicted_results}.

\subsection{Variability in Conversion Analysis Results Across Model Instances}\label{app:variability}
\begin{figure*}[t]
    \centering
    \includegraphics[height=5in]{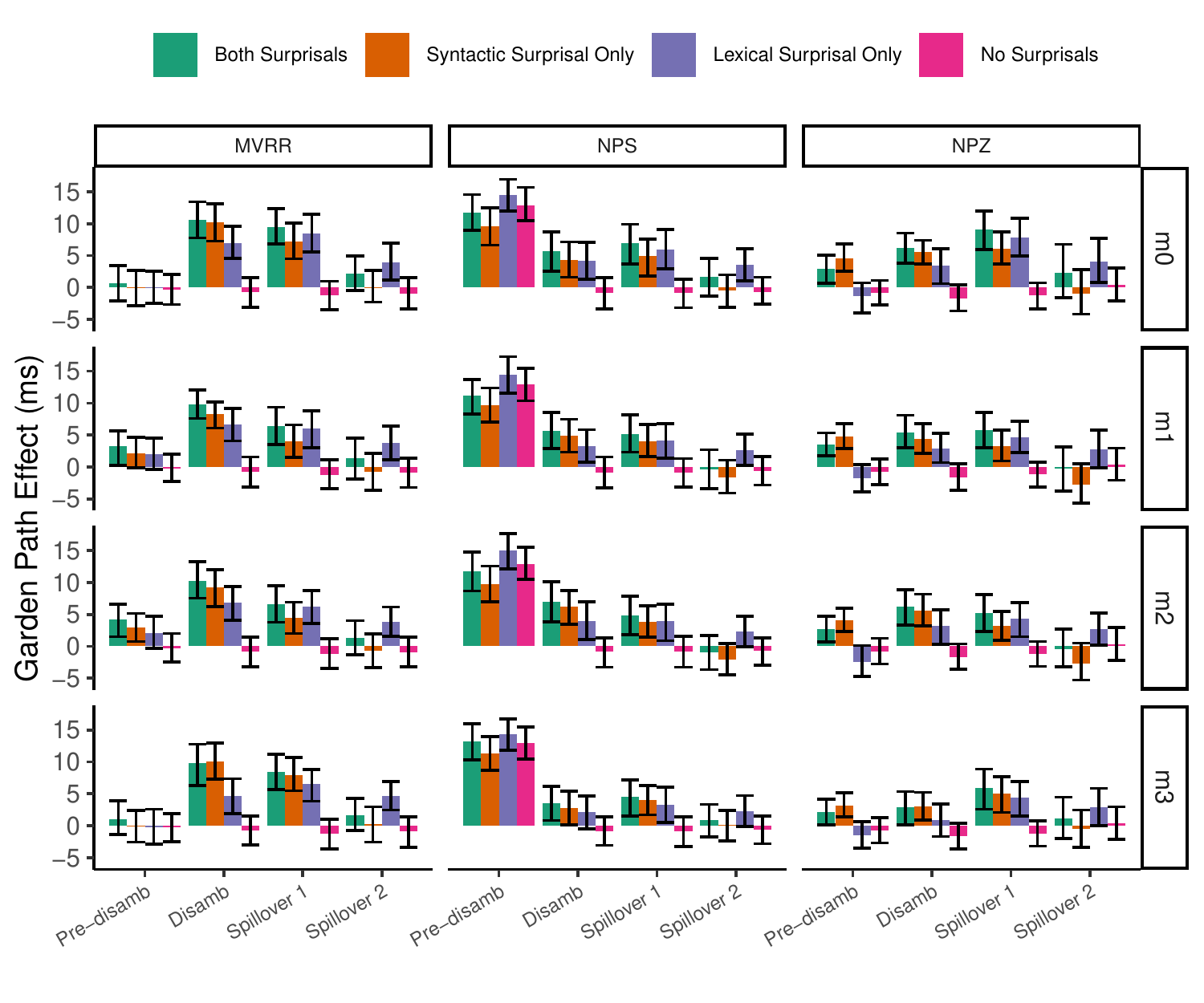}

    \caption{Empirical and model-predicted readings times for the three garden path constructions, broken out by the LM/Supertagger models used to generate the surprisals. Bars indicate the difference between the mean readings times for the ambiguous and unambiguous sentences across participants for each condition. Error bars indicate bootstrapped 95\% confidence intervals.}
    \label{fig:results_by_model}
    
\end{figure*}

In order to assess the robustness of our results with respect to the randomness in the training of our neural network models, we repeated our analysis using surprisals generated from four instances of our LM/supertagging model. These models differed only in the random seed used during the initialization and training procedure. In Figure~\ref{fig:results_model} in the main text, we presented predicted reading times averaged across these analyses. In Figure~\ref{fig:results_by_model} we present the same results broken out across each model instance.
\end{document}